\def\year{2021}\relax
\documentclass[letterpaper]{article} % DO NOT CHANGE THIS
\usepackage{aaai21}  % DO NOT CHANGE THIS
\usepackage{times}  % DO NOT CHANGE THIS
\usepackage{helvet} % DO NOT CHANGE THIS
\usepackage{courier}  % DO NOT CHANGE THIS
\usepackage[hyphens]{url}  % DO NOT CHANGE THIS
\usepackage{graphicx} % DO NOT CHANGE THIS
\usepackage{natbib}  % DO NOT CHANGE THIS AND DO NOT ADD ANY OPTIONS TO IT
\usepackage{caption} % DO NOT CHANGE THIS AND DO NOT ADD ANY OPTIONS TO IT
\usepackage[switch]{lineno}  %
\usepackage{nicefrac}
\usepackage{dsfont}

\newtheorem{theorem}{Theorem}%[section]

\newtheorem{question}[theorem]{Research Questions}

\usepackage{color}
\usepackage[textsize=scriptsize]{todonotes}
\definecolor{blue}{RGB}{0, 93, 170}			%Go Big Blue!
\definecolor{darkgreen}{RGB}{0, 102, 0}
\definecolor{orange}{RGB}{255, 160, 0} %'Cuse is in the house

\newcommand{\ignore}[1]{}
\usepackage{xspace}

\title{Thinking Fast and Slow in AI}
\author{
    G. Booch, \textsuperscript{\rm 1}
    F. Fabiano, \textsuperscript{\rm 2}
    L. Horesh, \textsuperscript{\rm 1}
    K. Kate, \textsuperscript{\rm 1}
    J. Lenchner, \textsuperscript{\rm 1}
    N. Linck, \textsuperscript{\rm 1}
    A. Loreggia,\textsuperscript{\rm 3} \\
    K. Murugesan, \textsuperscript{\rm 1}
    N. Mattei, \textsuperscript{\rm 4},
    F. Rossi, \textsuperscript{\rm 1}
    B. Srivastava \textsuperscript{\rm 5} \\
    }
\affiliations{
    \textsuperscript{\rm 1} IBM \\
    \textsuperscript{\rm 2} University of Udine \\
    \textsuperscript{\rm 3} European University Institute \\
    \textsuperscript{\rm 4} Tulane University \\
    \textsuperscript{\rm 5} Universty of South Carolina
}
\begin{document}

\maketitle

%\begin{center} {\bf \Large Thinking Fast and Slow in AI}\\
%G. Booch (IBM), F. Fabiano (Univ. Udine), L. Horesh (IBM), K. Kate (IBM), J. Lenchner (IBM), N. Linck (IBM), A. Loreggia (EUI), K. Murugesan (IBM), N. Mattei (Tulane Univ.), F. Rossi (IBM), B. Srivastava (USC)
%\end{center}

\begin{abstract}
This paper proposes a research direction to advance AI which draws inspiration from cognitive theories of human decision making. The premise is that if we gain insights about the causes of some human capabilities that are still lacking in AI (for instance, adaptability, generalizability, common sense, and causal reasoning), we may obtain similar capabilities in an AI system by embedding these causal components. We hope that the high-level description of our vision included in this paper, as well as the several research questions that we propose to consider, can stimulate the AI research community to define, try and evaluate new methodologies, frameworks, and evaluation metrics, in the spirit of achieving a better understanding of both human and machine intelligence.
%These presentations aim at presenting ideas and visions that can stimulate the research community to pursue new directions, e.g., new problems, new application domains, or new methodologies that are likely to stimulate significant new research. The presenter should find the right arguments to convince the audience that the topic is promising, and should relate the talk as much as possible to the existing literature.

%In AI, the ability to model and reason with preferences allows for more personalized services. Ethical priorities are also essential, if we want AI systems to make decisions that are ethically acceptable.  Both data-driven and symbolic methods can be used to model preferences and ethical priorities, and to combine them in the same system, as two agents that need to cooperate. We describe two approaches to design AI systems that can reason with both preferences and ethical priorities. We then generalize this setting to follow Kahneman's theory of thinking fast and slow in the human's mind. According to this theory, we make decision by employing and combining two very different systems: one accounts for intuition and immediate but imprecise actions, while the other one models correct and complex logical reasoning. We discuss how such two systems could possibly be exploited and adapted to design machines that allow for both data-driven and logical reasoning, and exhibit degrees of personalized and ethically acceptable behavior.
\end{abstract}

\section{Motivation and Overall Vision}

%We are motivated by the limitations of the current AI technologies.

AI systems have seen dramatic advancement in recent years, bringing many successful applications that are pervading our everyday life. However, we are still mostly seeing instances of narrow AI:
each of these developments are typically focused on a very limited set of competences and goals, e.g., image interpretation, natural language processing, label classification, prediction, and many others. Moreover, while these successes can be accredited to improved algorithms and techniques, they are also tightly linked to the availability of huge datasets and computational power \cite{marcus2020next}. State-of-the-art AI still lacks many capabilities that would naturally be included in a notion of intelligence, for example, if we compare these AI technologies to what human beings are able to do. Examples of these capabilities are generalizability, robustness, explainability, causal analysis, abstraction, common sense reasoning, ethics reasoning, as well as a complex and seamless integration of learning and reasoning supported by both implicit and explicit knowledge.

The majority of the AI community is currently making several attempts to address the current limitations of AI and create systems that display the ability for more human-like qualities, using a variety of approaches. One of the main debates is whether end-to-end neural network approaches can achieve this goal? or whether we need to integrate machine learning with symbolic and logic-based AI techniques? We believe that the integration route is the most promising, and this is supported by several results that have been obtained along this line of work. For example, Bengio conjectures that it is necessary to generalize from raw data to a "consciousness stream" of few a concepts sparsely related to each other
%Bengio has a proposal -- Position paper about building a consciousness stream out of the data stream
 \cite{bengio2017consciousness}. Also, Marcus argues that explicit knowledge, symbols, and reasoning should be used to improve the robustness of current AI systems
\cite{marcus2020next}.
Other researchers are building hybrid systems that use both machine learning and symbolic reasoning techniques, employing a so-called neuro-symbolic AI approach \cite{neuro-symbolic}.
%An example is AlphaGo, that exploits both machine learning and Monte-Carlo tree search to excel at playing Go.
%For example,
%The Neuro-Symbolic Concept Learner: Interpreting Scenes, Words, and Sentences From Natural Supervision
%\cite{mao2018neuro}

%Thinking, fast and slow: Combining vector spaces and knowledge graphs
%\cite{mittal2017thinking}

%Thinking fast and slow: Optimization decomposition across timescales
% \cite{goel2017thinking}

%Deep reasoning networks: Thinking fast and slow
%\cite{chen2019deep}

%Thinking fast and slow with deep learning and tree search
%\cite{anthony2017thinking}

We argue that a better comprehension of how humans have, and have evolved to obtain, these advanced capabilities can inspire innovative ways to imbue AI systems with these competencies. We propose to study and exploit cognitive theories of human reasoning and decision making as a source of inspiration for the causal source of these capabilities, that help us raise the fundamental research questions to be considered when trying to provide AI with desired dimensions of human intelligence that are currently lacking.

More precisely, we analyze some of these theories, with special focus on Kahneman's theory of thinking fast and slow, and attempt to connect them into a unified theory with the aim to identify (some of) the roots of the desired human capabilities. We then propose to translate these theories of the human mind into the AI environment, and we conjecture that this will possibly lead to some of the same kind of capabilities in machines.
This paper gives a brief and high-level overview of this general approach, tries to motivate it, lists several research questions along the way, and ends with a call for action to AI researchers to convene an AI research community that explores this space, assesses the validity of this approach, and possibly uses it to make significant advances in AI as well as in understanding human intelligence.

\section{Thinking Fast and Slow, and Other Theories of Human Decision Making}

According to Daniel Kahneman's theory, described in his book ``Thinking, Fast and Slow" \cite{kahneman2011thinking},
humans' decisions are supported and guided by the cooperation of two main kinds of capabilities, that, for sake of simplicity are called ``systems": system 1 provides tools for intuitive, imprecise, fast, and often unconscious decisions (``thinking fast"), while
system 2 handles more complex situations where logical and rational thinking is needed to reach a complex decision
(``thinking slow").

%Humans have the ability to adapt to very different scenarios. We can face situations in which making decision demands for a fast and intuitive choice. But we can also experience more intricate situations where we are challenged to reason and deliberate about an action to take.
%The decision making process has been described as a cooperation between two systems or agents that interact together to make a decision \cite{kahneman2011thinking}.
System 1 is guided mainly by intuition rather than deliberation. It gives fast answers to very simple questions. Such answers are sometimes wrong, mainly because of unconscious bias or because they rely on heuristics or other short cuts, and usually do not have an explanation.
However, system 1 is able to build models of the world that, although inaccurate and imprecise, can fill the knowledge gaps through causal inference and allow us to respond reasonably well to the many stimuli of our everyday life.
A typical example of a task handled by system 1 is finding the answer to a very simple arithmetic calculation, or reaching out to grab something that is going to fall. We use our system 1 about 95\% of the time when we need to make a decision.
%. Right because guided by intuition, these answers do not need any elaborated process to be computed, sometimes they could also be wrong and they do not necessarily have a specific explanation.

When the problem is too complex for system 1, system 2 kicks in and solves it with access to additional computational resources, full attention, and sophisticated logical reasoning. A typical example of a problem handled by system 2 is solving a complex arithmetic calculation, or a multi-criteria optimization problem.
To do this, humans need to be able to recognize that a problem goes beyond a threshold of cognitive ease and therefore the need to activate a more global and accurate reasoning machinery. Hence, introspection is essential in this process.
%System 2 is activated when a more elaborated answer is required and the first process is not able to provide an answer.

When a problem is new and difficult to solve, it is handled by system 2. However, certain problems over time pass on to system 1. The reason is that the procedures used by system 2 to find solutions to such problems are used to accumulate examples that system 1 can later use readily with little effort. Thus, after a while, some problems, initially solvable only by resorting to the system 2 reasoning tools, can become manageable by system 1. A typical example is reading text in our own native language. However, this does not happen with all tasks. An example of a problem that never passes to system 1 is finding the correct solution to complex arithmetic questions (unless one is a mathematical genius).

While system 2 seems capable of solving harder problems than system 1, system 2 does not usually work by itself, but it is rather supported by system 1 in its elaborate calculations. When searching for a solution in a very large solution space, system 2 does not usually explore the whole search space but may employ heuristics that are provided by system 1 and that help in focusing the attention only on the most promising parts of the space. This allows system 2 to work with manageable time and space.

System 1 is also able to perform some basic forms of causal reasoning. This allows it to build a (possibly imprecise and biased) model of the world even if it has incomplete knowledge, and to use this model to tackle simple tasks. The system 1 data-level causality skills also support the more complex and accurate reasoning of system 2 on problems which are cognitively more complex.

With this complex internal machinery, humans can reason at various levels of abstraction, adapt to new environments, generalize from specific experiences to reuse their skills in other problems, and can easily multi-task when using their system 1. On the other hand, system 2 is sequential, since it requires full attention to devise and execute the appropriate solving procedure for a complex or new problem. Thus only one complex task at a time can be handled by a human being.

Another way to look at this division is that system 1 works at a local level, activating only a specific part of the brain, e.g., when we recognize a familiar face, or when we speak. On the other hand, system 2 involves more regions of the brain and combines their contributions, so it works at a more global level.
It is therefore the presence of multiple specialized agents/components/skills that supports the functioning of both systems. We recall however that system 1 and system 2 are not systems in the usual multi-agent architecture terminology, but rather they are metaphors and short-hands for two broad classes of information processing.

While Kahneman's theory gives a detailed account on how we make decisions, the theory of Y. N. Harari \cite{harari2014sapiens} conjectures what led us to have such reasoning capabilities over generations, and why our decision making engine came to have this organization.
%While Another theory which is related to Kahneman's one is Harari's theory of the history of humans .
According to Harari's theory, Homo-Sapiens had greater success than animals or other primates and human species because over (evolutionary) time
it acquired the ability to conceive and communicate high-level stories. These skills  allowed Sapiens to collaborate flexibly with others at scale, and thus to have a broader impact and a larger influence upon the world. This capability was supported by the ability to abstract from raw data to high level concepts, that could then be communicated more easily and compactly. As noted above, this is primarily a system 2 capability, since it requires a voluntary and conscious decision that goes beyond an involuntary reactive behavior (and its communications to others) to be able to abstract, generalize, and reason about scenarios and concepts that do not physically exist in real life (the so-called ``stories" in Harari's terminology).
The ability to reason at different levels of abstraction requires also another mechanistic capability: to focus attention to certain, limited set of features and process them in depth, while disregarding others, since deferring (temporarily) some of the features of a concept is the essence of an abstraction process \cite{zucker2003grounded}.

While these theories are drawing hypotheses as to the causes of human behavior and decisions, due to the current incomplete knowledge and understanding of our brain, they are supported by ample evidence and experimental results.
%For example, further substantiation to Harari's theory can be found in both evolutionary and developmental analysis of the neural system. Functional mapping of brain regions (as performed by MacLean \cite{maclean1990triune} in his triune brain and by Kanwisher in her work on functional neuro-anatomical mapping \cite{kanwisher1997fusiform}) enables contrasting the  reptilian complex with the paleomammalian cortex and the neo-cortex, and establishes that several functionalities attributed to system 2 (e.g. causal planning and abstraction) has evolved anatomically late in the evolution.

The notion of consciousness is important as well, when trying to identify the traits of human (or animal) intelligence. According to M. Graziano \cite{graziano2013consciousness,graziano2020toward}, there are two forms of consciousness in human beings: the I-consciousness (I for Information) and the M-consciousness (M for mysterious).
The first one refers to the ability to solve (possibly complex) problems, by recognizing necessary processing steps
in specific (even new) context, to tackle a desired problem.
This conduct seems to be related to system 2's high-fidelity processing mode, since it has to do with considering a problem and harnessing the relevant faculties of our cognition to devise a plan to solve it.
The latter refers to our ability to build a simplified, approximate, and subjective model of peoples', both ourselves and others, mind, beliefs and intentions.
%the mind itself and of others. This form of consciousness enables an approximate, subjective reconstruction of content related to peoples mind, beliefs and intentions. S
Such low-fidelity model building can be linked to system 1, as system 1 is able to form a rapid but usually inexact model of the world.
While many animals and primates possess various levels of M-consciousness, and also limited forms of I-consciousness, humans excel at I-consciousness.
M-consciousness can  even sustain in presence of limited I-consciousness, yet, a sophisticated I-consciousness needs to rely on M-consciousness: without a model of the mind (of self and other agents), it is difficult to devise how to adapt to new environments and solve complex problems.

\section{AI Thinking, Fast and Slow}

The theories described in the previous section, as well as their connections, shed some light on which competencies provide humans the ability to solve a diverse set of simple and complex problems; understand broad contexts robustly; adapt readily at short time scales (rather than evolutionary time scales) to new scenarios and environmental conditions; and ultimately cooperate at scale.
These competences include abstraction, generalization, causal analysis and planning, attention, locality, skills combination, introspection, and various forms of implicit and explicit knowledge.

We now discuss some of these concepts and capabilities in the AI context, with the aim to identify central research questions that could help understand how to equip AI with the capabilities that it still lacks.
The intent is to stimulate the AI research community to collectively study such questions, possibly raise other ones, and find appropriate answers.
%to discuss how to embed them and  to possibly obtain In other words, it seems that important capabilities comprising the product of human decision making abilities are still lacking in AI. We therefore believe it is worthwhile to study how to embed (some form of) these capabilities into AI, suitable adapted to a different platform, with the aim to achieve similar abilities.
%While we do not have specific solutions,
%the goal of this paper is mainly to define and discuss some research questions around how to equip AI with these capabilities, with the hope that the AI research community will collectively study such questions, possibly raise other ones, and find appropriate answers.

%So we propose to rely on them to build a general theory of machine reasoning and decision making that can provide machines with the capabilities that can support adaptability, robustness, and cooperation.
%What they tell us is that humans exploit this dual-system machinery to employ certain essential human capabilities, including generalizability and adaptability, that machine still largely lack.
%Humans also know how to use efficiently their conscious reasoning, which would force them to be sequential, by recognizing when a task requires a careful, high level, and correct response. \nick{From call today: need to reiterate that System 1 and System 2 are metaphores -- do we need to define what they are for a machine? -- System1 and 2 describe broad categories of capabilities and involve many components.}

\subsection{System 1 and System 2 in AI}

It is possible to draw a very loose parallel between the capabilities included in system 1 and system 2, and the two main lines of work in AI: machine learning (ML) and symbolic logic reasoning, or data-driven vs knowledge-driven AI. System 1 seems to have similar traits as ML, with its ability to build (possibly imprecise and biased) models from sensory data. For example, perception activities, such as seeing and reading, that are currently addressed with ML techniques, are usually handled by human's system 1.
However, system 1 exploits its ability to grasp basic notions of causality to build such models, and this does not seem to be present, at least for now, in ML. Another system 1 trait that is lacking in ML is common-sense reasoning (that of course is also related to causality).
On the other hand, system 2's capability to solve complex problems seems to be related to AI techniques based on logic, search, optimization, and planning, and employing explicit knowledge, symbols, and high-level concepts.

However, as noted earlier, these two broad sets of capabilities are usually symbiotic, which supports hybrid neuro-symbolic approaches in AI.

%Given the description of the capabilities of System 1/System 2 in humans, system 1 capabilities are similar to machine learning techniques.
%For example, perception activities, that are now addressed with ML techniques, are usually handled by human's system 1.
%On the other hand, complex reasoning processes, such as those employed in planning and scheduling, are handled by system 2.

%Mention existing work by Bengio on injecting system 2 capabilities in DL.

\begin{question}
Should we clearly identify the AI system 1 and system 2 capabilities? What would their features be? Should there be two sets of capabilities, or more?
%How to define the AI system 1 capabilities?
%what are their features?
%Should they be techniques that are mostly greedy and miopic? And what about the AI system 2? Should it include instead techniques that are guided by look-ahead and exploration?
\end{question}

While the human system 2 is sequential, and its activation requires our full attention, it does not necessarily have to be this way in a machine.
%How to define the AI system 1 capabilities? what are their features? Should they be techniques that are mostly greedy and miopic? And what about the AI system 2? Should it include instead techniques that are guided by look-ahead and exploration?

\begin{question}
Is the  sequentiality  of system 2 a bug or a feature? Should we carry it over to machines or should we exploit parallel threads performing system 2 reasoning? Would this, together with the greater computing power of machines compared to humans, compensate for the lack of other capabilities in AI?
\end{question}

%\subsubsection{Holistic Evaluation of AI Systems}
Consider an AI system that is based on the two main sets of capabilities provided by ML and symbolic AI. Usually
%A practical issue with regard to system building is how to evaluate how good the system is doing. The
the quality of ML approaches is measured by the degree to which they can achieve the desired result, e.g., accuracy, precision, recall, F1 score. On the other hand, the quality of a symbolic reasoning process is usually measured by the correctness of its conclusions, e.g., the satisfaction of a set of constraints. In a combined system1/system2 AI system which switches dynamically between the two opportunistically, or combines their capabilities, selecting evaluation measures and methods is an open research question. See \cite{nlp-testing} for a recent framework to evaluate NLP systems regardless of task and implementation.

\begin{question}
What are the metrics to evaluate the quality of a hybrid system 1/ system 2 (or ML/ symbolic) AI system?
Should these metrics be different for different tasks and combination approaches?
\end{question}

\subsection{Introspection and Governance}

%\subsubsection{Introspection and Conscience.}

%According to M. Graziano \cite{graziano2013consciousness,graziano2020toward}, there are two forms of consciousness in human beings: the I-consciousness (I for Information) and the M-consciousness (M for mysterious).
%The first one refers to the mechanistic ability to solve problems, by recognizing what needs to be done in each specific context. This system can be related primarily to system 2 modes of operations, it is the specialization capability to attend to a specific mission and harness the required faculties of our cognition to attend to such goals.

%The latter, refers to our ability to build simplified model of the mind itself and of others. This form of consciousness, enables an  approximate, subjective reconstruction of content related to peoples mind, beliefs and intentions. Such, low-fidelity model of processing, can be linked to Kahneman's system 1, with its unique ability to form a rapid, yet, inexact model of the world.

%explain about machines, and how I and M consciousness in machine

Our aim is to build machines that are able to autonomously figure out how to utilize their cognitive resources to solve any challenge that comes their way. To achieve this, a subjective (if imprecise) model of agents' minds, such as the one underlying the concept of M-consciousness, is probably needed in order to identify the relevant factors for assessing the features of the problem.  For example, if the scenario is cognitively easy or difficult, if it is risky to make a mistake, if solving it gives a high reward, etc., as well as to assess the competencies of the mechanistic system's components (the I-consciousness) to solve the problem, and select the most appropriate solution procedure.

\begin{question}
How do we define AI's introspection in terms of I-consciousness and M-consciousness?
Can M-consciousness alone solve complex problems that in humans need also sophisticated levels of I-consciousness?
How is introspection linked to autonomy and human-machine teaming?
Is introspection a binary concept or should it have several levels? If in levels, how can we associate different capabilities to the various levels?
Should we have models of the agents' minds (those supporting M-consciousness) at different levels of precision/fidelity?
\end{question}

%nd account for both the cost of performing the task for our own mind, and the respective reward associated with successful processing.

% add research question

%What do we want introspection to be based on ?

%there are attributes of M and I consciousness ... overlap

%This is the kind of consciousness that we would call ``introspection" in a machine: TBD

%competency aware

%link to multi-fidelity studies

%present as a conjecture

%\subsubsection{Switching between System 1 and System 2.}

Through introspection, an AI system should be able to recognize when it needs to deploy capabilities from system 1 or system 2, and also when to switch between the two systems. %how does it know when to employ the more demanding System 2 capabilities?
In humans, such a switch depends on many factors,  including framing, priming, and the context of the decision environment. In an AI system, there seems to be other factors that make up the sufficient conditions for this switch, e.g., when system 1 cannot find a solution.
%. For example, when system 1 cannot find a solution, e.g., because we are in a new environment, or when there are several alternatives
%divergent policies/solutions
%that must be considered before choosing an action.
We may also need to switch when we are facing a very high stake decision: a mistake may confer dangerous consequences according to some high-weight criterion. Or also when the decision needs to be explainable to a third party.

%Given the above behavioral causes, other question arise naturally:
%\begin{itemize}
    %\item How do we define divergence or absence of solutions?
    %\item How do we recognize the presence of these conditions?
    %\item How should system 2 act once the switch is triggered?
%\end{itemize}

\begin{question}
How do we model the governance of system 1 and system 2 in an AI? When do we switch or combine them? Which factors trigger the switch? How do we define the divergence between or absence of potential solving procedures?
%How to recognize the presence of these conditions?
How should system 2 act once the switch is triggered?
%When should problems, that until then are handled by system 2, be instead deferred to be solved by system 1 hereafter?
When should problems be handed back from system 2 to system 1?
How to interject the notion of time in the governance framework?
\end{question}

\subsection{Divergence Resolution and Ethical Reasoning}

As noted above, the notion of divergence is important for the governance framework. But it is also fundamentally tied to system 2 capabilities. When there are two or more diverging policies or heuristics offered up by system 1, and it is important to solve the problem correctly, one needs to carefully explore the main pros and cons, as well as the possible consequences of our actions, before deciding on the policy to adopt, or we need to define a new policy. This careful reasoning, based on inference and counter-factuals, seems to be a system 2 capability, supported by system 1's heuristics to reduce the search space. We have investigated how to incorporate possibly divergent value functions, say the tension between immediate reward and long term constraints imposed by behavioral norms in a variety of reinforcement learning settings \cite{balakrishnan2018incorporating,BaBoMaRo18,DBLP:journals/ibmrd/BalakrishnanBMR19} and probabilistic planning \cite{NBMC19+,DBLP:journals/ibmrd/NoothigattuBMCM19}.

\begin{question}
How can we leverage a model based on system 1 and system 2 in AI to understand and reason in complex environments when we have competing priorities?
Which part of the AI system recognizes the divergence and, supported by the introspection, identifies the best solving procedure? What should be done existing solvers are not able to solve a given problem?
\end{question}

Complex constraints and competing objectives generate dilemmas that humans confront every day \cite{RoMa19a}.
A typical example is when we need to make a decision that has some ethical component, such as deciding if a certain action is morally acceptable or not.
Usually we simulate various approaches to the given question, based for example on deontology or consequentialism \cite{kagan2018normative}, and then we make our moral decision.
%The aforementioned theories and approaches suggest how the interleaving between the two systems drive humans in making a decision.
Preliminary work along these lines attempts to understand how humans switch between these different approaches
in judging the morality of an action \cite{rossi2019preferences,Awad:BreakRules},
how divergence can be defined in an ethical context \cite{loreggia2018}, and how to define methodologies for ethical reasoning in AI \cite{LoMaRoVe18,LoMaRoVe18a,loreggia2019cpm}.

\begin{question}
Which capabilities are needed to perform various forms of moral judging and decision making?
Do they require a specific notion of divergence?
How do we model and deploy possibly conflicting normative ethical theories in AI?
Are the various ethics theories tied to either system 1 or system 2?
%or do they need a combination of the two systems?
\end{question}

\subsection{Abstraction, Generalization, and Knowledge}

%Let us consider humans' capability to generalize and adapt. %We put them together since they are tightly connected.
In order to adapt to a new environment, an agent needs to be aware of its competencies (e.g., introspection considered earlier) and know how to deploy its skills and problem solving capabilities, were possibly acquired in another environment, into the new one. This process is supported by the ability to abstract and generalize the specific skills and their intervention target object, in order to subsequently specialize them again in the new scenario. So, adaptability involves first an abstraction/generalization phase and then an instantiation phase.
Of course both these steps are not random, but are guided by the ability to recognize the similarities and differences between the two environments, and to use this knowledge to decide what to temporarily forget during the abstraction step. This conjecture is aligned with the so-called consciousness prior theory \cite{bengio2017consciousness}. Existing and well established abstraction theories \cite{cousot}
and studies of various forms of abstraction in AI \cite{zucker2003grounded}
can be useful, coupled with attention mechanisms \cite{attention}.

\begin{question}
How do we define abstraction / generalization mechanisms that are guided by a notion of attention and pass from the raw data level to a more abstract level? How do we know what to forget from the input data during the abstraction step?
Should we keep knowledge at various levels of abstraction, or just raw data and fully explicit high-level knowledge?
What does it mean for knowledge to be explicit: is it related to the presence of metadata, structured knowledge graphs, or language-related entities?
\end{question}

\subsection{Multi-agent Environment, Epistemic Reasoning, and Architectural Support}

Humans live in societies, and their individual decisions are linked to their perception of reality, which includes the world, the other agents, and themselves. These models are not perfect but they are used to make informed decisions, as well as a test bed to evaluate consequences of alternative decisions. However,
such models are not based on exact knowledge, but rather
on approximate information on the world and on beliefs on what others know and believe. Multi-agent epistemic reasoning~\cite{fagin2003reasoning} is what humans engage on when making complex decisions using system 2, unless such decisions are completely independent of other agents or their consequences are irrelevant.

\begin{question}
In a multi-agent view of several AI systems communicating and learning from each other, how to exploit/adapt current results on epistemic reasoning and planning to build/learn models of the world and of others?
\end{question}

Following Misky's theory of the society of mind \cite{minsky}, as well as Brooks' subsumption architecture approach \cite{brooks1990elephants}, we believe that AI systems should
include several independent simple components, to be triggered when needed according to a governance model.
This implies that the best architectural support for this vision of AI is a multi-agent architecture where the individual agents focus on specific skills and problems, act asynchronously, contribute to building models of the world, of other AI systems, and of self, and can be combined in many ways.

\begin{question}
What architectural choices best support the above vision of the future of AI?
\end{question}

{
%\small
%\bibliographystyle{aaai21.bst}
\bibliography{abb,thinking_aaai21}
}

\end{document}